\documentclass[11pt, letterpaper]{cmu}

\usepackage[comma,numbers,sort,compress]{natbib}
\usepackage{microtype}
\usepackage{float}
\usepackage{mathtools}
\usepackage{mathrsfs}
\usepackage{nicefrac}
\usepackage{dsfont}
\usepackage{bbm}
\usepackage{multirow}
\usepackage[export]{adjustbox}
\usepackage{wrapfig}
\usepackage{subcaption}
\usepackage{dblfloatfix}
\usepackage{listings}
\usepackage{setspace}
\usepackage{algpseudocode}
\usepackage{algorithm}
\usepackage{array}
\usepackage{makecell}
\usepackage{lineno}
\usepackage{soul}
\usepackage[most,skins,theorems]{tcolorbox}
\usepackage[textsize=tiny]{todonotes}
\usepackage[capitalize,noabbrev]{cleveref}
\usepackage{capt-of}  
\usepackage{etoolbox}

\hypersetup{
    colorlinks = true,
    citecolor = {magenta},
    linkcolor = {blue},
    urlcolor = {blue},
}

\definecolor{highlightmistake}{RGB}{255, 179, 179}
\definecolor{highlightcorrect}{RGB}{179, 255, 179}
\definecolor{rliableolive}{HTML}{BBCC33}
\definecolor{rliableblue}{HTML}{77AADD}
\definecolor{rliablered}{HTML}{EE8866}
\definecolor{myblue}{rgb}{0.1,0.4,0.9}
\definecolor{lightblue}{rgb}{0.22,0.45,0.70}
\definecolor{violet}{HTML}{6A0DAD}
\definecolor{darkblue}{rgb}{0, 0, 0.5}

\newcolumntype{Y}{>{\centering\arraybackslash}X}
\newcolumntype{L}[1]{>{\raggedright\arraybackslash}p{#1}}

\newtcolorbox{promptbox}[2][]{
listing only,
enhanced,
breakable,
colback=rliableolive!13!white,
colframe=black,
fontupper=\ttfamily,
title=#2,
#1}

\tcbset{
  aibox/.style={
    width=\linewidth,
    top=8pt,
    bottom=4pt,
    colback=rliableolive!8!white,
    colframe=black,
    colbacktitle=black,
    enhanced,
    center,
    attach boxed title to top left={yshift=-0.1in,xshift=0.15in},
    boxed title style={boxrule=0pt,colframe=white,},
  }
}
\newtcolorbox{AIbox}[2][]{aibox,title=#2,#1}

\newcommand{\methodname}{{ExpRL}}

\newcommand{\ourtitle}{\methodname{}: Exploratory RL for LLM Mid-Training}

\usepackage{amsmath,amsfonts,bm}

\def\eqref#1{Eq.~\ref{#1}}

\def\1{\bm{1}}

\DeclareMathAlphabet{\mathsfit}{\encodingdefault}{\sfdefault}{m}{sl}
\SetMathAlphabet{\mathsfit}{bold}{\encodingdefault}{\sfdefault}{bx}{n}

\newcommand{\pibase}{\pi_b}

\newcommand{\by}{\mathbf{y}}

\newcommand{\bx}{\mathbf{x}}

\title{\ourtitle}

\author[1]{Violet Xiang}
\author[2]{Amrith Setlur}
\author[3,$\dagger$]{Chase Blagden}
\author[1]{Nick Haber}
\author[2]{Aviral Kumar}

\affil[1]{Stanford University}
\affil[2]{Carnegie Mellon University}
\affil[3]{OpenAI}
\affil[$\dagger$]{Work done while at Rogo.}

\correspondingauthor{ziyxiang@stanford.edu}

\providecommand{\theabstract}{}

\begin{document}
\maketitle

\vspace{-1.5cm}
\begin{center}
\begin{minipage}{0.94\textwidth}
  \centering
  \includegraphics[width=0.90\textwidth]{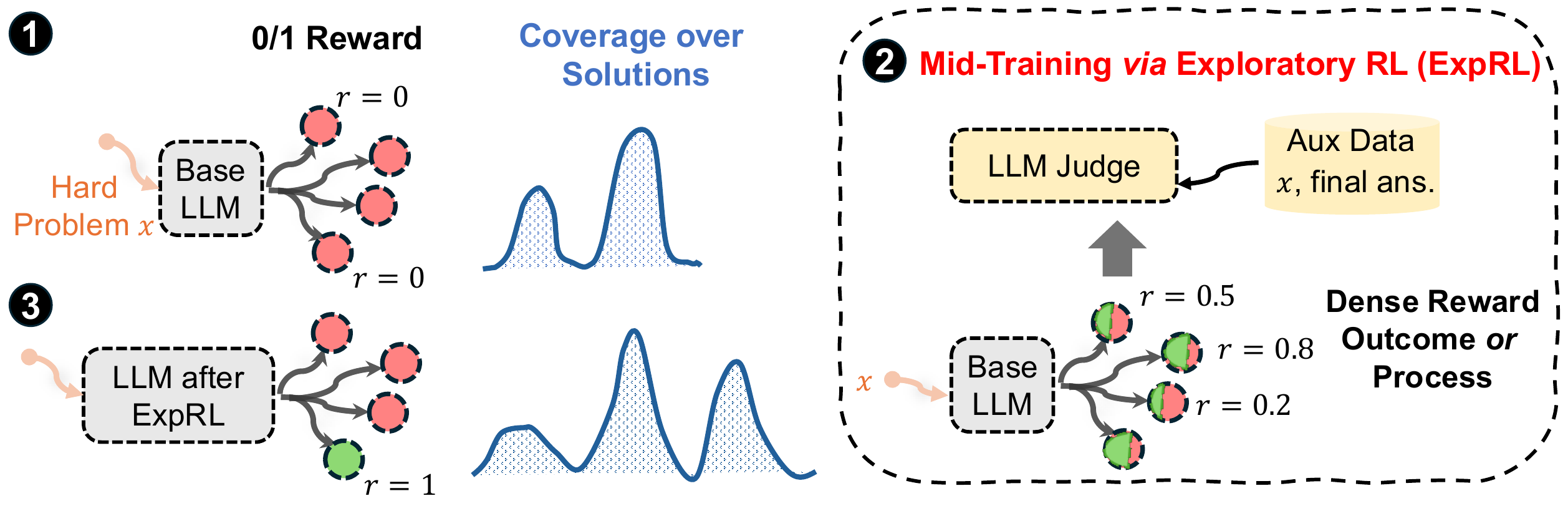}
  \vspace{-0.15cm}
  \captionof{figure}{\footnotesize{\textbf{Exploratory RL (ExpRL).}
  (1) On hard problems we fail to achieve high outcome-level correctness
  (sparse rewards) under the base LLM since it lacks coverage over diverse
  solutions needed to solve problems. (2) To build coverage, we mid-train
  the base LLM with our approach exploratory RL or ExpRL. In particular,
  we use unstructured auxiliary information (reference solutions on hard
  problems) to reward partial progress made by the base LLM via
  \methodname{}-Outcome and \methodname{}-Process rewards given by an LLM judge,
  and run RL with these rewards. (3) \methodname{} is able to achieve
  non-trivial correctness, as judged by sparse outcome-level correctness,
  making it a well-primed initialization for subsequent sparse-reward RL.}}
  \label{fig:schema}
\end{minipage}
\end{center}
\vspace{-0.35cm}

\noindent\textbf{Abstract: }Sparse reward reinforcement learning (RL) has become a standard tool for improving LLM reasoning, but its success depends critically on the coverage present in the base model. In practice, models are often primed for RL through \emph{mid-training} on curated reasoning traces that teach useful primitive skills such as decomposition, verification, or self-correction. Although effective, this strategy requires manually specifying what the model should learn, and it remains unclear whether such primitive coverage is enough for much harder problems, which require combining these skills into broader solution strategies. We study a more automated approach: \emph{RL-based mid-training} using large corpora of human-written question-answer data. Rather than treating reference solutions as targets to imitate, our method, \methodname{}, uses them as \emph{reward scaffolds}: references are hidden from the policy and used only to construct problem-specific grading rubrics for judging on-policy reasoning traces.
The policy samples from the original problem prompt, while an LLM judge compares the sampled reasoning trace against the reference solution and assigns outcome-level or process-level dense rewards. This lets \methodname{} reinforce partial progress, useful intermediate reductions, and productive reasoning behaviors that sparse final-answer rewards often fail to upweight. On challenging math reasoning tasks, \methodname{} yields stronger RL priming than SFT, sparse-reward GRPO, and self-distillation, and provides a better initialization for subsequent sparse-reward RL. Additional mixed-domain experiments further suggest that \methodname{} can extend beyond the original math-only setting.

\vspace{0.3em}
\noindent\textbf{Code:} \url{https://github.com/violetxi/ExpRL}
\vspace{0.3em}

\vspace{-0.4cm}
\section{Introduction}
\vspace{-0.2cm}
Reinforcement learning (RL) has become a standard tool for improving the reasoning abilities of large language models (LLMs). Yet its success depends critically on the \textbf{coverage} present in the base model before RL begins. If the base model assigns very little probability to useful reasoning paths, then even many sampled attempts may produce few correct or partially correct trajectories. In this regime, sparse final-answer rewards give little signal to learn from, and RL mainly reinforces behaviors the model already samples well. Initialization is therefore a central bottleneck for scaling RL-based reasoning.

The goal of mid-training is to improve this starting distribution before sparse-reward RL. Operationally, we want the model to place more probability mass on productive reasoning attempts across a wide range of hard problems, so that pass@$k$ improves and downstream RL has more useful trajectories to reinforce. This coverage is shaped both by primitive reasoning skills, such as decomposition, verification, backtracking, and self-correction~\citep{gandhi2025cognitivebehaviorsenableselfimproving}, and by the model's ability to compose these skills into broader problem-solving techniques. For example, knowing how to check local computations does not mean the model can identify the right case split for a hard combinatorics problem and carry it through. Our aim is to build this broader coverage: not merely to improve correctness on the mid-training distribution, but to upweight productive reasoning paths that make later sparse-reward RL more effective. Operationally, we view pass@$k$ as an observable proxy for this coverage. If a policy assigns nontrivial probability mass to at least one complete productive reasoning path for a problem, then repeated sampling should eventually uncover a correct rollout. Thus, improvements in pass@$k$ indicate that mid-training has expanded the set of solution strategies the model can sample, even when pass@1 remains limited.

In this work, we study how to use reference solutions for RL-based mid-training rather than imitation-based mid-training. When such reference solutions are available, a natural baseline is to convert them into supervised question-solution traces and fine-tune the model to imitate them. However, directly cloning traces that rely on solutions unlikely under the base model can disrupt its reasoning abilities~\citep{yang2026int,kang2024unfamiliar}. Another baseline in this reference-solution setting is on-policy self-distillation, as explored in~\citep{hubotter2026reinforcement}. This reduces the off-policy mismatch from SFT by training on the model's own rollouts, but the supervision signal still comes from token-level targets induced by a privileged teacher. When this target distribution is far from what the student can reliably produce, such supervision may hurt generalization~\citep{kim2026does}. Thus, imitation and distillation are natural baselines when reference solutions are available, but they may be limited mechanisms for broadening coverage over productive reasoning paths.

We therefore propose \textbf{Exploratory RL} (\textbf{\methodname{}}), an RL-based mid-training method that uses reference solutions to provide dense rewards for on-policy reasoning traces. Rather than exposing references as demonstrations or hints, \methodname{} uses them as reward scaffolds: it helps the judge construct a problem-specific rubric for scoring partial progress in the actor's on-policy rollout. Because the policy samples only from the original problem prompt, this preserves on-policy exploration while providing richer feedback than the typical  sparse outcome-level correctness rewards.

Concretely, \methodname{} assigns each on-policy rollout a \emph{partial progress} score by comparing it with a reference solution for the same problem. We study two variants: \emph{\methodname{}-Outcome} assigns a dense outcome-level reward to the full rollout. \emph{\methodname{}-Process}  rewards intermediate prefixes with dense scores, giving  local credit to partial progress. These rewards can reinforce promising decompositions, correct intermediate reductions, or useful solution structures even when the model does not yet solve the problem fully. 
In this way, reference solutions help RL upweight productive reasoning paths without exposing the reference to the actor as a target trajectory or oracle prefix.

We evaluate \methodname{} in an RL-priming setting on challenging answer-based math reasoning. We compare against SFT, sparse-reward GRPO, and self-distillation. \methodname{} produces a stronger Stage-I policy and a better initialization for subsequent sparse-reward RL. The gains appear not only in pass@1 but also in pass@$k$ and in the diversity of reasoning attempts, consistent with improved coverage over productive reasoning paths. We also find that \methodname{} changes the model's reasoning behavior, increasing verification, self-correction, and backtracking relative to the base model. Finally, we test \methodname{} in a broader mixed-domain setting and analyze when reference-conditioned judging provides reliable rewards. These results suggest that \methodname{} can serve as a general priming interface using reference answers.

\vspace{-0.3cm}
\section{Preliminaries, Definitions, and Notation}
\label{sec:prelim}
\vspace{-0.2cm}

\textbf{Problem setup.}
We are given a large dataset $\mathcal{D}_\text{mid} = \{(\bx_i, \by_i^\star)\}_{i=1}^N$, where $\bx_i$ denotes a problem and $\by_i^\star$ is a step-by-step reference solution. Typically, these reference solutions are human-written and may differ substantially in style from LLM-generated reasoning traces.
We use $\pi_\theta$ to denote an LLM policy with trainable parameters $\theta$, and $\pibase$ to denote the base pre-trained LLM. \emph{Our goal} is to train $\pibase$ on $\mathcal{D}_\text{mid}$ so as to build broader coverage over productive reasoning paths that will help it subsequently solve problems from a downstream dataset $\mathcal{D}'$ (which may or may not be similar to $\mathcal{D}_\text{mid}$) when trained further with RL using only a sparse \emph{binary outcome reward} $r(\bx,\by)\in\{0,1\}$, indicating whether the rollout's final answer is correct (\textit{e.g.}, by string-matching a final boxed answer).

\textbf{Evaluating exploratory capabilities.} Our primary downstream metric is pass@1 after Stage-II sparse-reward RL on $\mathcal{D}'$, which measures reliable single-sample performance. We also report pass@$k$, the probability of sampling at least one correct rollout in $k$ independent attempts, as an operational proxy for coverage under sampling. Higher pass@$k$ indicates that the policy assigns more probability mass to reasoning paths that can lead to a correct solution. We measure pass@$k$ on $\mathcal{D}_\text{mid}$ to diagnose whether RL priming increases coverage where reference-guided rewards are applied, and on $\mathcal{D}'$ to test whether this exploratory capability transfers to downstream sparse-reward RL.

\textbf{RL priming for downstream RL.} We use \emph{RL priming} to refer to any (mid-)training procedure that prepares a base model for a downstream later RL stage with binary rewards. \textit{E.g.}, for math reasoning this would mean imbuing a base model with the ability to compose primitive skills into productive reasoning paths needed for downstream RL on hard problems. Ultimately, we say that a model is well primed for downstream RL if its pass@$k$ on $\mathcal{D}'$ is large.

\textbf{RL algorithms.} As we discuss shortly, our approach for RL priming will involve running RL with a dense reward signal applied at both outcome level (in the end) and \emph{process level} (intermediate points). For our RL runs with outcome rewards (dense or sparse), we use  GRPO~\citep{guo2025deepseek} with normalization applied across $n$ rollouts per problem. For our implementation of process rewards, we use the REINFORCE~\citep{ahmadian2024basicsrevisitingreinforcestyle} update. Here, we still sample $n$ rollouts per problem and compute the following gradient  for the current policy $\pi$ given a batch of problems $\bx_1, \ldots, \bx_N$, each with $n$  responses. The batch gradient $ \nabla_\pi J(\pi)$ is then given by:
\begin{align}
    \nabla_\pi J(\pi) := \frac{1}{N}\sum_{i \in [N]} \frac{1}{n}  \sum_{j \in [n]} \sum_{k \in [|\by_j|]} A(\bx_i, \mathbf{y^{*}_{i}}, y_i^{k}) \cdot \nabla_\pi  \log \pi (y_i^{k} \mid \by^{<k}_{i})
\end{align}
In the above expression $\pi(y_j^k\!\mid\!\mathbf{y}_j^{<k})$ is the probability of the $k^\mathrm{th}$ token in response $\by_j$. In particular,  note  that the advantage $A(\bx_i, y_j^k)$ 
is computed at the token level since rewards and advantages differ at different positions in a rollout, when using process rewards. Later we outline our construction of the advantage function for this setting.

\vspace{-0.3cm}
\section{\methodname: Reference-Guided Dense Rewards for RL Priming}
\label{sec:our_method}
\vspace{-0.2cm}
In this section, we introduce \methodname{}, an RL-based priming stage before downstream sparse outcome-reward RL. \methodname{} uses dense (non-binary) rewards at the outcome or process level derived from reference solutions on a broad mid-training dataset of question-answer pairs.
RL with these dense rewards is intended to broaden coverage over productive reasoning paths. Stage-I gains in pass@1 and pass@$k$ serve as diagnostics that the primed policy assigns more probability mass to trajectories that can reach correct solutions. The aim is not to hand-specify isolated skills, but to induce a broader repertoire of useful reasoning behaviors for subsequent sparse-reward RL to reinforce.
As reflected by the poor pass@$k$ for the base model (Qwen3-4B-Instruct) in Table~\ref{tab:stage1_pass}, it is clear that it lacks sufficient coverage over reasoning paths even though it may consist of useful reasoning behaviors. Since human-written solutions are substantially different from the model's own reasoning traces, directly acquiring such coverage through offline training (\textit{e.g.}, SFT) is challenging~\citep{yang2026int}, which motivates an on-policy RL procedure.

\textbf{\emph{Design principle for \methodname{}.}} Since the goal of \methodname{} is to broaden the model's coverage, a sparse outcome reward on final correctness for sampled on-policy rollouts is insufficient on questions in the mid-training data. It only indicates whether a rollout eventually reaches a correct answer, without discriminating between rollouts that make useful intermediate progress and those that do not. Our design principle is therefore to reward on-policy traces based on how likely they are to reach reference solutions for problems in the mid-training data, using a dense reward even when the overall trace is incorrect. As long as the mid-training dataset contains hard questions that require diverse reasoning patterns, this procedure should help shift probability mass toward more productive reasoning paths and provide a stronger initialization that covers useful reasoning paths for downstream RL. \methodname{} uses an LLM-based judge to measure similarity to reference solutions and instantiates this principle through both outcome-level and process-level rewards, as we discuss next.

\emph{\textbf{Concrete approach: RL priming via dense reference-guided rewards.}} Building on this principle, our approach uses reference solutions to construct dense rewards. Doing so is possible because modern LLMs are often better at verifying partial progress against a reference than at generating a correct solution from scratch. We exploit this verification-generation gap to assign informative scores, thereby ranking sampled traces by how much useful progress they exhibit. Optimizing these rewards shifts probability mass toward regions of the space of traces that are more likely to result in an eventual success, improving pass@$k$ and, more importantly, building a stronger exploration prior for subsequent sparse reward RL.

\textbf{Step I: Assigning numerical dense rewards via reference-guided verification.}
To obtain dense signals, we ask the model to compare self-generated solutions against provided reference solutions.
We instantiate our base model as our LLM judge $J$ that scores a candidate solution $\by$ by comparing it to the reference $\by^\star$ under a fixed rubric, which measures alignment between the generated trace and techniques or high-level strategies in the reference solution (see Appendix~\ref{sec:a_prompts} for the rubric).
Formally, given $(\bx, \by, \by^\star)$, the judge outputs a score $
\tilde{s}(\bx, \by, \by^\star) \in \{1,2,3,4,5\}$:
\begin{align*}
s(\bx, \by, \by^\star)
\;=\;
\frac{\tilde{s}(\bx, \by, \by^\star)-1}{4}
\;\in\; [0,1].
\end{align*}
The judge is explicitly instructed to \emph{verify rather than solve} and not to introduce missing steps, fill in unstated intermediate results, or correct errors in the model output. If a rubric item is not directly supported by the text of $\by$, it is scored as being absent. Because the entire reference solution $\by^\star$ is available, this comparison yields a dense learning signal even when rollouts with correct final answers are rarely sampled by the model on problems in the mid-training set. After generating these rewards, we use them downstream in two ways to instantiate ExpRL.

\textbf{a) \methodname{}-Outcome.}
Using the reference-guided score, we define an \methodname{}-Outcome reward on full traces sampled from the mid-training data: $s(\bx, \by, \by^\star)$.
Unlike sparse outcome rewards, this provides graded feedback to partially correct solutions that match the reference under the rubric but fail later, preserving distinctions among unsuccessful rollouts and providing useful signal even when fully correct solutions are rarely sampled. These rewards are used only during exploratory mid-training and need not perfectly reflect task success, as long as they encourage exploration over productive reasoning paths.

\textbf{b) \methodname{}-Process.} While outcome-level dense rewards provide a more frequent learning  signal compared to sparse outcome, they do not localize credit within the sampled rollout. To improve credit assignment, we also consider a process-level reward from partial rollouts, i.e., rollout prefixes. Given a generated solution $\by$, we form a sequence of prefixes $\{\by_{\le t}\}_{t=1}^T$ according to a fixed rule for slicing prefixes (that we discuss in~\ref{sec:slice_steps}), and apply the same judge to each prefix to obtain:
\begin{align*}
s_t
\;=\;
s(\bx, \by_{\le t}, \by^\star),
\qquad
t = 1,\ldots,T.
\end{align*}
These prefix scores provide intermediate feedback about partial progress toward the reference solution. Intuitively, process-level rewards encourage early decisions that are predictive of eventual success, while avoiding over-crediting prefixes that later degrade.

\textbf{Process-level advantage normalization.}
Although $\{s_t\}_{t=1}^T$ provides absolute judge scores for each prefix, we convert them into \emph{centered} segment-level advantages to emphasize \emph{relative} partial progress rather than absolute score calibration across problems. Specifically, we use
\begin{equation}
A_t(x,y)=
\begin{cases}
s_t - s_{t-1}, & \text{if } t > 1,\\
s_1 - s_T,     & \text{if } t = 1.
\end{cases}
\label{eq:pr_delta}
\end{equation}
For $t>1$, a segment receives positive advantage only if it improves judged alignment with the reference relative to the previous prefix, and negative advantage if it reflects regression. This encourages the policy to build on intermediate progress. We center the first segment as $A_1 = s_1 - s_T$ rather than using $A_1 = s_1$ directly, so that its scale is comparable to later differences and the first step does not dominate the update. We use this $A_t$ as the process-level learning signal in the on-policy objective below.

\textbf{Step II: Optimization objective and training details.}
We optimize the policy using on-policy RL with KL regularization against a reference policy $\pi_0$:
\begin{align}
\max_\theta \;
&\mathbb{E}_{(\bx,\by^\star)\sim\mathcal{D}_\text{mid}} \left[
\mathbb{E}_{\by\sim\pi_\theta(\cdot\mid\bx)}
\big[ R(\bx,\by,\by^\star) \big] -
\beta\,\mathrm{KL}\!\left(\pi_\theta(\cdot\mid\bx)\,\|\,\pi_0(\cdot\mid\bx)\right) \right],
\label{eq:rl_objective}
\end{align}
where $R$ denotes the outcome or process-level dense rewards defined above. In particular, for \methodname{}-Outcome rewards, we use $s(\bx, \by, \by^*)$ directly and apply a GRPO-style update that normalizes scores across the batch. For \methodname{}-Process rewards, we instead substitute the advantages from Equation~\ref{eq:pr_delta} into the GRPO update,
and do not apply any other normalization.
Since this stage is used only for mid-training, the rewards need not be perfectly accurate; they only need to encourage broad and diverse behaviors. We ablate alternative normalization strategies for process advantages in Appendix~\ref{sec:advantage_ablation}.

\textbf{Running downstream RL after \methodname{}.}
After mid-training with \methodname{}, we initialize downstream RL from the primed policy and train on the target dataset $\mathcal{D}'$ using the standard sparse outcome reward. The downstream objective and reward structure are unchanged; only the initialization differs. In implementation, we use two different on-policy RL pipelines (details in \ref{sec:additional_imp}) for Stage 1 and Stage 2, replacing the reference-guided dense rewards used by \methodname{} with binary final-answer rewards during downstream RL. By shifting probability mass toward productive reasoning trajectories before sparse-reward training begins, \methodname{} increases the likelihood that downstream RL encounters informative rollouts early in training.

\begin{AIbox}{Summary: Mid-Training \textit{via} Exploratory RL or ExpRL}
ExpRL is an online RL approach for mid-training that uses reference solutions to define \emph{dense} outcome/process rewards rather than traces to imitate. By rewarding partial progress rather than pure correctness, it shifts probability mass toward productive reasoning paths, building stronger coverage for later \emph{sparse} \mbox{reward} RL.
\end{AIbox}

\vspace{-0.3cm}
\section{Experiments}
\label{sec:experiments}
\vspace{-0.2cm}

The goal of our experiments is to evaluate the efficacy of \methodname{} for improving the base model for subsequent RL training. To this end, we run \methodname{} on a dataset of challenging math question-answer pairs that the base model fails to solve in 64 independent samples, each with a 32k-token response budget. We compare \methodname{} against alternative mid-training procedures on this same data, then run Stage-II sparse-reward RL from each resulting initialization and evaluate on held-out math benchmarks. We describe our setup next and then present our results. 

\vspace{-0.2cm}
\subsection{Setup: Dataset, Evaluation Protocol, and Training Hyperparameters}
\label{sec:task_dataset_eval}
\vspace{-0.1cm}

\textbf{Base model, judge, and training datasets for \methodname{}.}
We use Qwen3-4B-Instruct-2507 as the policy backbone (Qwen3-4B-Instruct for brevity). This model is trained to produce reasoning traces directly within the chain of thought, without needing a `$<$think$>$' block. We produce dense rewards using an LLM judge based on the same Qwen3-4B-Instruct model akin to \citet{yang2026int}. In the main experiments, we use a copy of the base model as the judge; Sec~\ref{sec:mixed_domain_calibration} shows that ExpRL can also work when a smaller reference-conditioned judge provides rewards for a larger policy. We train the model to optimize Eq.~\ref{eq:rl_objective} via REINFORCE. More specifically, sparse-reward baselines and Stage-II use GRPO-style group normalization. \methodname{}-Outcome uses a GRPO-style normalized reward update. \methodname{}-Process uses REINFORCE-style token/segment advantages without group normalization. For the prompts, we use a dataset combining hard question and reference answer pairs from recent works \textbf{InT}~\citep{yang2026int} and \textbf{POPE}~\citep{qu2025pope}.

\textbf{Mid-training for RL priming (Stage-I).}
Unless otherwise stated, we sample $G{=}10$ rollouts per prompt with temperature $0.8$ and a maximum generation length of $16{,}384$ tokens during training. This length budget is rarely reached by the initial policy, but provides headroom for RL-induced length growth without aggressively truncating longer reasoning traces. We assign the entire trajectory a reward of $0$ when a generation overflows the maximum length. This prevents degenerate training dynamics in which the policy can increase reward by producing overly long outputs. For producing \methodname{}-Process rewards in \methodname{}, given a rollout $\by$, we define a sequence of segment prefixes $\{\by_{\le t}\}_{t=1}^T$ using the delimiter \texttt{"\#\#\#"}; we use this delimiter because the model defaults to emitting it between reasoning steps. We query the judge on each prefix to obtain $s_t = s(\bx, \by_{\le t}, \by^\star)$, where $s_T$ denotes the score of the full rollout, and compute the segment-level advantages described in Equation~\ref{eq:pr_delta} for all $t < T$. These advantages are then used as process-level learning signals at the segment level. We train the \methodname{} stage for $230$ optimization steps. For the downstream final-answer-reward RL stage, we train for $500$ optimization steps. Unless otherwise specified, we use a per-update prompt batch size of $36$ for runs using the LLM judge and $32$ for verifiable sparse reward RL runs. More details in Appendix~\ref{sec:additional_imp}.

\textbf{Downstream RL (Stage-II).}
After each Stage-I procedure, we use the resulting policy as the initialization for downstream sparse-reward GRPO. In the main math experiments, Stage-II training uses the InT+POPE prompt mixture, the same prompt family used during \methodname{} priming, but with all reference-solution information removed: the policy samples from the original problem prompt and receives only the binary final-answer reward. This is the same-distribution instance of our problem setup, where $\mathcal{D}'$ is drawn from the same problem family as $\mathcal{D}_\text{mid}$. The experiment tests whether reference-guided priming produces an initialization that makes ordinary sparse-reward RL more effective.

\textbf{Benchmarks.} We consider four standard, held-out answer-based reasoning benchmarks: \textbf{HMMT (November 2025)}, \textbf{IMO-AnswerBench}~\citep{luong2025towards},
\textbf{AIME 2025}, and \textbf{AIME 2026}. We sample 128 responses per problem to compute evaluation metrics. In all cases, we evaluate both the base model initialization obtained after training with \methodname{} as well as the model obtained after downstream RL training.

\begin{table*}[b]
\centering
\small
\setlength{\tabcolsep}{6pt}
\renewcommand{\arraystretch}{1.08}
\begin{tabular}{lcccc}
\toprule
\textbf{Method}
& \textbf{AIME25} $\uparrow$
& \textbf{AIME26} $\uparrow$
& \textbf{HMMT} $\uparrow$
& \textbf{IMO Answer} $\uparrow$ \\
\midrule
Qwen3-4B-Instruct     & 46.46 & 51.40 & 40.60 & 31.37 \\
SFT                 & 26.62 & 30.26 & 20.09 & 21.80 \\
GRPO                & 55.99 & 58.75 & 42.91 & 35.28 \\
Self-Distillation   & 55.59 & 58.41 & 46.08 & 35.18 \\
\midrule
\methodname{}-Outcome (Ours)       & \textbf{59.07} & 61.74 & \textbf{49.11} & \textbf{37.85} \\
\methodname{}-Process (Ours)      & 58.08 & \textbf{63.41} & 48.13 & 35.73 \\
\bottomrule
\end{tabular}
\caption{\footnotesize{\textbf{Pass@1 on answer-based benchmarks after downstream sparse-reward RL.}
Models are initialized with different RL-priming methods and then continued with the same Stage-II RL setup, except Qwen3-4B-Instruct as the original base model.
Generally speaking, ExpRL attains the strongest overall answer-based performance.}}
\label{tab:stage2_answer}
\end{table*}

\vspace{-0.3cm}
\subsection{Baselines and Comparisons}
\label{sec:baselines}
\vspace{-0.2cm}

We compare ExpRL to several approaches, including \textbf{(1) SFT:} supervised fine-tuning on the reference solutions in the mid-training set instead of running RL for mid-training; \textbf{(2) Verifiable sparse reward RL:} standard GRPO using only a binary final-answer reward from a rule-based verifier on mid-training prompts, without any dense reference-guided feedback; and \textbf{(3) Self-distillation:} a distillation-based baseline in which sampled rollouts are trained against a richer self-teacher; concretely, we use the base model conditioned on the reference solution as the teacher following \citet{hubotter2026reinforcement}.

We compare these baselines against two variants of \methodname{}: a) \textbf{\methodname{}-Outcome}, which assigns a dense terminal reward to full rollouts, and b) \textbf{\methodname{}-Process}, which assigns dense rewards to partial rollouts and prefixes. 
The key distinction is that the distillation baselines use reference information to define token-level targets, whereas \methodname{} uses the same information only to score sampled reasoning traces and shape the model's exploration prior before subsequent sparse-reward RL. 
We focus our comparisons on methods that use reference information during the priming stage while leaving downstream sparse-reward RL unchanged. Prefix-guided methods such as \textbf{POPE}~\citep{qu2025pope} study a complementary setting: they guide exploration by exposing oracle prefixes during downstream RL. In contrast, \methodname{} uses references only to construct rewards during priming, and can in principle be combined with prefix-guided exploration.

\vspace{-0.3cm}
\subsection{Finding 1: \methodname{} Yields A Stronger Initialization for Downstream RL}
\vspace{-0.2cm}
Our main question is whether dense rewards in \methodname{} can produce a better initialization for downstream sparse-reward RL compared to imitation-based or sparse-reward-only alternatives. Table~\ref{tab:stage2_answer} shows that the answer is indeed \emph{yes}. After a stage of standard sparse reward RL, we observe that overall \methodname{} variants outperform SFT, sparse GRPO and self-distillation on the held-out answer-based benchmarks. The clearest gain appears on AIME-2026, where \methodname{}-Process 
reaches 63\% after downstream RL, while the second best GRPO baseline attains 58.75\% only. Across the remaining answer-based benchmarks, the \methodname{} variants consistently occupy the top of the table, with \methodname{}-Outcome strongest on multiple evaluations and \methodname{}-Process remaining highly competitive throughout. Taken together, these results support our central claim: for RL priming, privileged reference solutions are more effective when used to score sampled reasoning than when used only as trajectories to imitate.

\vspace{-0.3cm}
\subsection{Finding 2: \methodname{} Improves the Primed Policy Before Downstream RL}
\vspace{-0.2cm}

\begin{wrapfigure}{h}{0.4\textwidth}
    \centering
    \vspace{-0.3cm}
    \includegraphics[width=0.41\textwidth]{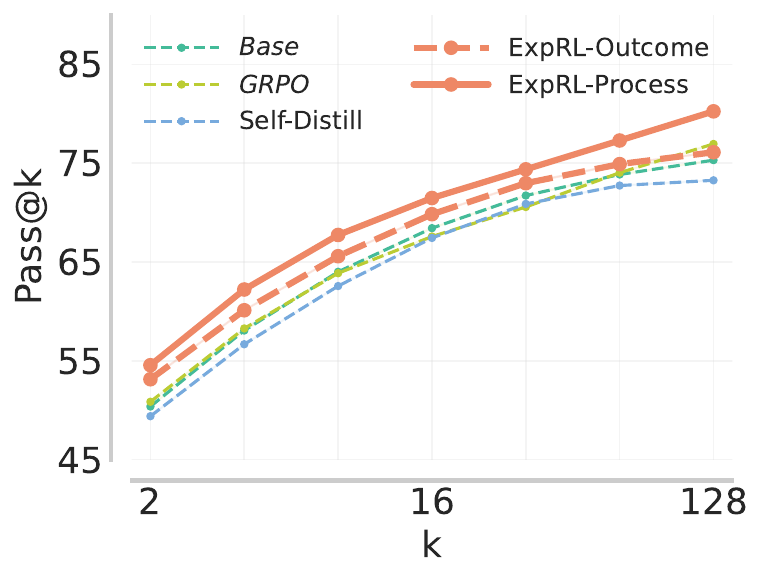}
    \vspace{-0.8cm}
    \caption{\footnotesize{\textbf{Pass@k after training} with \methodname{} on HMMT-Nov-2025 (128 samples).}}
    \label{fig:hmmt_passk}
    \vspace{-0.5cm}
\end{wrapfigure}

We next study if \methodname{}, in and of itself, is able to already improve performance of models after RL priming, even before running any downstream sparse reward RL. Observe in Table~\ref{tab:stage1_pass} \methodname{} already produces a stronger model on held-out answer-based benchmarks, achieving higher pass@1 and pass@$k$ and in cases such as IMO-AnswerBench it can improve the pass@$k$ coverage that downstream sparse-reward RL can subsequently amplify. 
As one representative example, Figure~\ref{fig:hmmt_passk} shows the \texttt{pass@k} curves on HMMT, where both \methodname{} variants achieve higher pass rates at low values of $k$, and the \methodname{}-Process variant remains particularly strong even at higher $k$ (see Appendix~\ref{sec:pass@k} for pass@$k$ curves on other benchmarks). This distinction is important for interpreting the gains after downstream sparse-reward RL (from the result discussed above in Finding \#1). If RL priming were only improving mid-training rewards, it would not necessarily translate into stronger downstream sparse-reward RL. Instead, \methodname{} improves both pass@1 and pass@k before the second stage even begins, indicating that it produces a better RL-ready initialization. The stage-II improvements in Table~\ref{tab:stage2_answer} are therefore consistent with the view that downstream RL is benefiting from a stronger starting policy, rather than merely from additional optimization.

\begin{table}[h]
\centering
\footnotesize
\setlength{\tabcolsep}{3.5pt}
\renewcommand{\arraystretch}{1.05}
\begin{tabularx}{\columnwidth}{>{\raggedright\arraybackslash}p{2.5cm}*{8}{>{\centering\arraybackslash}X}}
\toprule
\textbf{Method}
& \multicolumn{2}{c}{\textbf{AIME25} $\uparrow$}
& \multicolumn{2}{c}{\textbf{AIME26} $\uparrow$}
& \multicolumn{2}{c}{\textbf{HMMT} $\uparrow$}
& \multicolumn{2}{c}{\textbf{IMO Answer} $\uparrow$} \\
\cmidrule(lr){2-3} \cmidrule(lr){4-5} \cmidrule(lr){6-7} \cmidrule(lr){8-9}
& \textbf{pass@1} & \textbf{pass@16}
& \textbf{pass@1} & \textbf{pass@16}
& \textbf{pass@1} & \textbf{pass@16}
& \textbf{pass@1} & \textbf{pass@16} \\
\midrule
\!\!\!Qwen3-4B-Instruct   & 46.46 & 72.32 & 51.45 & 80.30 & 40.60 & 68.43 & 31.37 & 52.74 \\
SFT               & 6.00  & 30.95 & 5.68 & 34.24 & 3.41 & 23.91 & 4.22 & 31.07 \\
GRPO              & 48.67 & 76.37 & 51.39 & 77.55 & 41.68 & 67.58 & \textbf{34.35} & 54.58 \\
Self-Distillation & 42.98 & 71.39 & 53.91 & 78.32 & 39.89 & 67.44 & 30.46 & 52.62 \\
\specialrule{0.08em}{0.2em}{0.2em}
\multicolumn{9}{l}{\textbf{ExpRL (Ours)}} \\
\methodname{}-Outcome     & 50.52 & \textbf{77.25} & 57.45 & 81.04 & 44.19 & 69.84 & 33.56 & \textbf{55.73} \\
\methodname{}-Process     & \textbf{51.77} & 74.29 & \textbf{57.51} & \textbf{81.10} & \textbf{45.24} & \textbf{71.48} & 32.02 & 54.29 \\
\bottomrule
\end{tabularx}
\vspace{-0.3cm}
\caption{\footnotesize{\textbf{Pass@1 and Pass@16 after Stage-I (\methodname{} mid-training).} \methodname{} generally improves Stage-I pass@1 and pass@16 over the base and baselines, with the largest gains on AIME26 and HMMT. On IMO-AnswerBench, results are more mixed, although ExpRL-Outcome still achieves the best pass@16.}}
\label{tab:stage1_pass}
\end{table}

\begin{figure*}[h]
  \centering
  \includegraphics[width=0.97\textwidth]{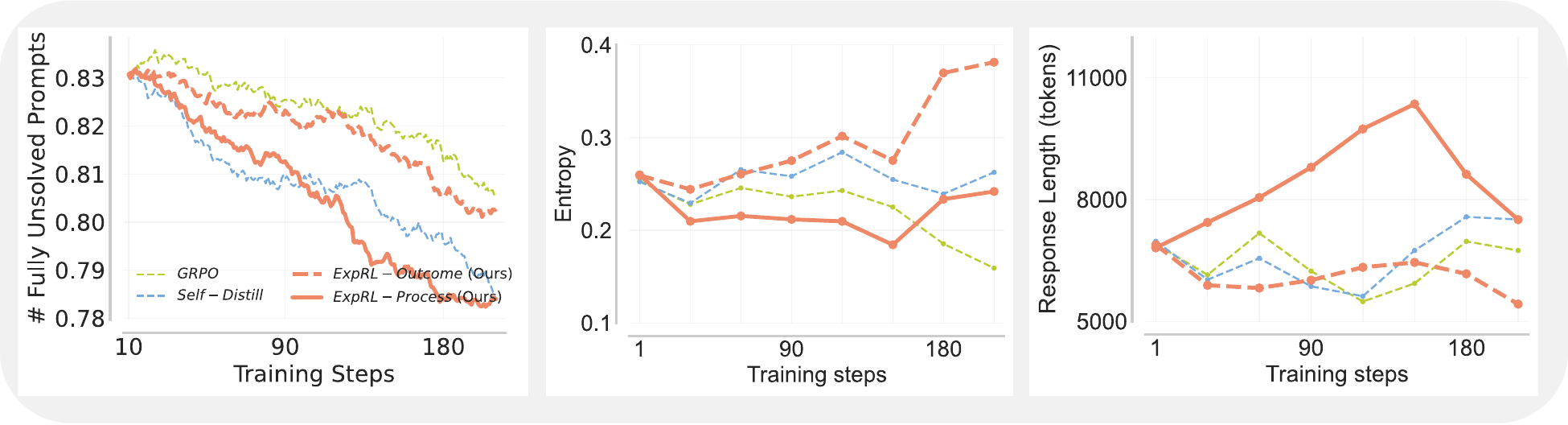}
  \caption{\footnotesize{\textbf{\methodname{} training dynamics during Stage-I.} \textbf{Left}: Number of unsolvable prompts (as measured by outcome-level correctness) reduces significantly faster in ExpRL with \methodname{}-Process rewards. \textbf{Middle}: token-level entropy remains relatively stable or even slightly increases for ExpRL and self-distillation, while it drops more substantially for sparse-reward GRPO. \textbf{Right}: Response length remains stable for all priming methods except ExpRL with process rewards which increases stably before reducing sharply when the responses get clipped.}}
  \label{fig:stage1_training_dynamics}
\end{figure*}

Figure~\ref{fig:stage1_training_dynamics} provides a complementary view of the training dynamics during RL priming. Entropy of sparse-reward RL (GRPO) collapses the fastest among the online methods and unlocks the fewest prompts over the course of training. In contrast, \methodname{} variants and self-distillation all maintain substantially higher token-level entropy, with \methodname{}-Process unlocking solvable prompts the fastest. We also observe distinct dynamics within the two \methodname{} variants. \methodname{}-Outcome shows a noticeable late increase in entropy without a similarly large increase in response length, whereas \methodname{}-Process exhibits an increase in response length. While these quantities are not themselves optimization targets, they suggest that \methodname{} does not simply sharpen the policy in the mode-seeking manner of sparse-reward GRPO, but instead alters the training dynamics in a way that is more consistent with building broader coverage in RL priming.

\begin{figure*}[h]
  \centering  
  \includegraphics[width=0.8\textwidth]{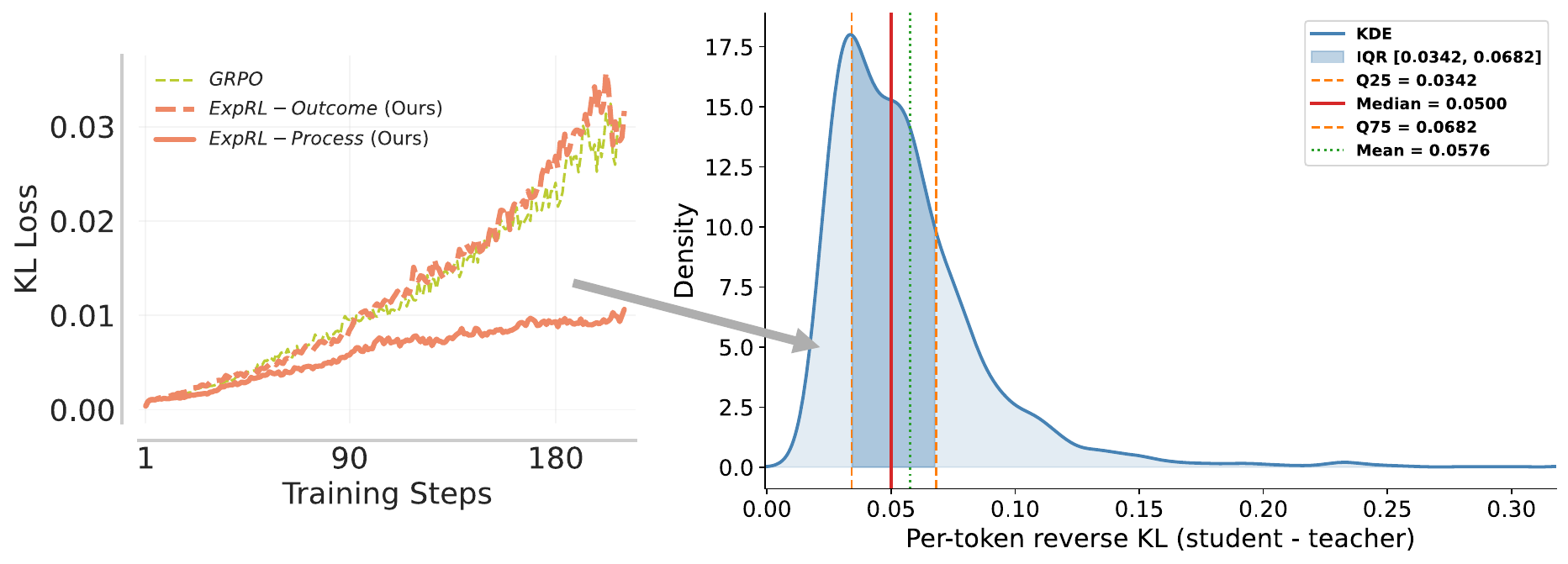}
  \vspace{-0.3cm}
  \caption{\footnotesize{\textbf{Teacher, $\pi_\text{teacher}$ used in self-distillation is far from the base model, $\pi_\text{student}$, in KL divergence.} \textbf{Left}: $\textrm{KL}[\pi_{\theta}||\pi_\text{ref}]$ to the reference policy during Stage-I training for GRPO and ExpRL. 
  \textbf{Right}: $\textrm{KL}[\pi_\text{student}||\pi_\text{teacher}]$ per problem $\bx$ at the start of self-distillation. We see that teacher is much farther outside the KL ball of policies reachable with on-policy reward optimization.}}
  \label{fig:self_distill_kl}
  \vspace{-0.3cm}
\end{figure*}

Figure~\ref{fig:self_distill_kl} provides an additional reason why self-distillation is a weaker RL-priming objective. At the start of self-distillation (Figure~\ref{fig:self_distill_kl} Right), the teacher starts much farther from the base model in KL than the other on-policy methods. This means that it begins from a substantially more off-policy target.  As prior work on off-policy distillation has noted that forcing a learner to match a distant expert distribution can lead to substantial distribution shift and unstable optimization~\citep{kang2024learning,setlur2026reuse}. In contrast, \methodname{} improves coverage while remaining within a more reachable KL space of the base policy.

\vspace{-0.3cm}
\subsection{Finding 3: \methodname{} Changes Reasoning Behaviors Relative to the Base LLM}

\vspace{-0.2cm}
\begin{figure*}[t]
  \centering
  \includegraphics[width=0.9\textwidth]{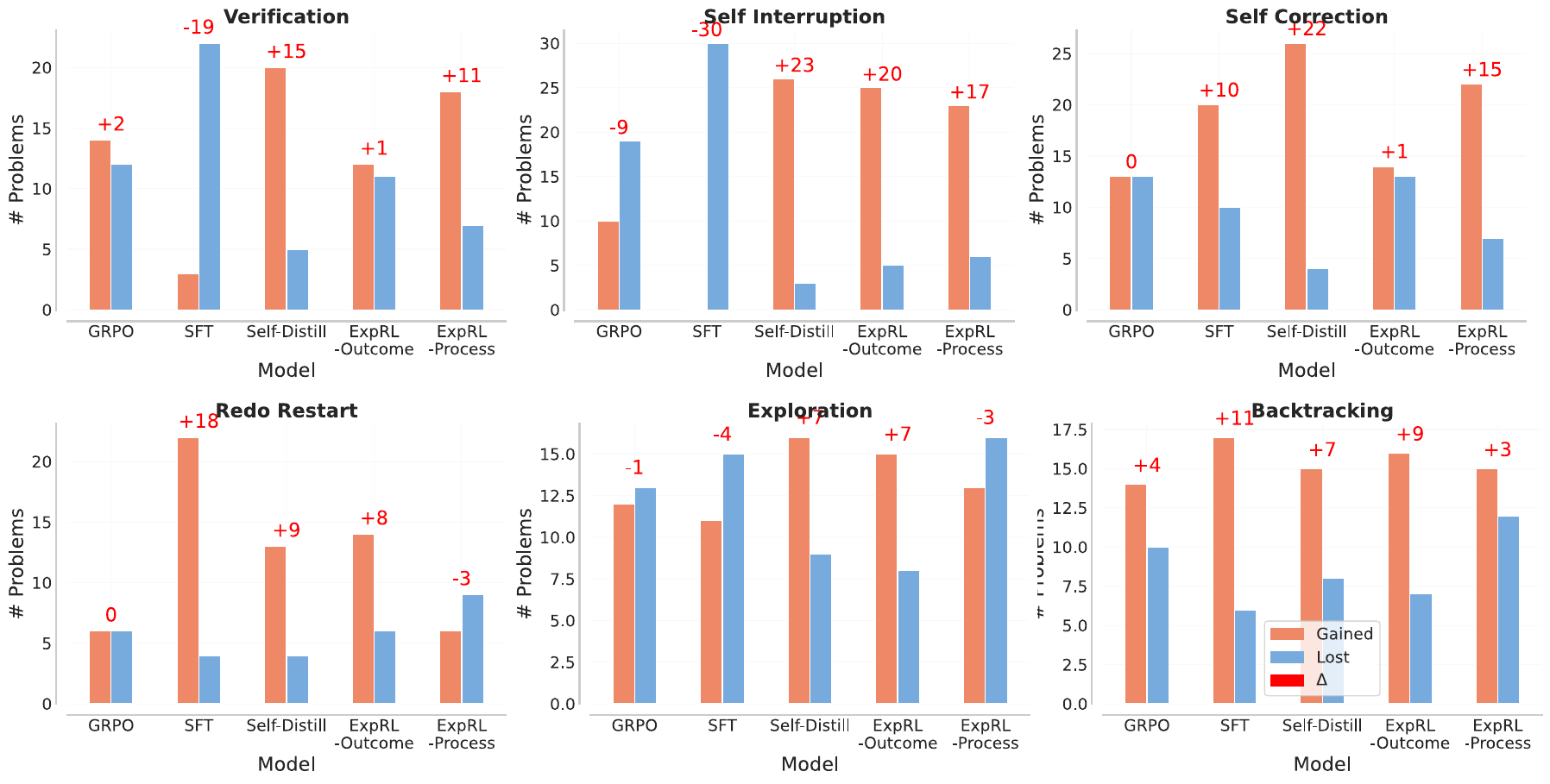}  
  \caption{\footnotesize{\textbf{Behavior changes after RL priming relative to the base model.} Orange bars show behaviors gained after priming, blue bars show behaviors lost, and red numbers indicate the net change. Search-oriented behaviors refer to observable rollout features in our annotation rubric, such as verification, self-correction, exploration, restarts, and backtracking. \methodname{} yields net gains in several such behaviors, suggesting that reference-guided RL priming changes the distribution of sampled trajectories rather than merely increasing final correctness.}}
  \label{fig:reasoning_behavior}
  \vspace{-0.55cm}
\end{figure*}

Beyond aggregate benchmark performance, we also ask whether \methodname{} changes the \emph{kind} of reasoning the model tends to produce (Section~\ref{sec:stage1_behavior}). Figure~\ref{fig:reasoning_behavior} suggests that it does. Relative to the base model, \methodname{} increases coverage over several of these search-oriented behaviors, especially verification, self-correction, and backtracking. Compared to SFT, which loses verification behavior, and sparse-reward GRPO, whose behavioral changes are smaller, \methodname{} better preserves or expands behaviors associated with adaptive search and sustained intermediate progress. Self-distillation provides an important contrast: it also increases several search-oriented behaviors, indicating that behavioral coverage alone is not unique to \methodname{}. However, its Stage-I pass@1/pass@$k$ and downstream sparse-RL performance are generally weaker than \methodname{} in Tables~\ref{tab:stage2_answer} and~\ref{tab:stage1_pass}. This suggests that useful RL priming requires both forms of coverage: coverage over behaviors that help scale test-time compute, and coverage over problem-specific knowledge and productive solution paths. Taken together, the behavioral analysis supports the main interpretation of \methodname{}: reference-guided RL priming changes the distribution of sampled trajectories in a way that increases useful search behaviors while also improving coverage over productive solution paths.

\vspace{-0.3cm}
\subsection{Finding 4: \methodname{} Extends to Mixed-Domain Mid-Training and Smaller Judges}
\label{sec:mixed_domain_calibration}
\vspace{-0.2cm}

The preceding experiments focus on a 4B policy and primarily math reasoning. We next test two questions about the scope of \methodname{}. First, can reference-guided RL priming improve a larger policy on a broader mixture of domains? Second, is the dense reward signal actually tied to the problem-matched reference, or can it be explained by generic LLM-judge confidence? To answer these questions, we run an additional mixed-domain Stage-I experiment and a judge/reference calibration stress test.

\textbf{Mixed-domain scale study.}
We construct a Stage-I mixture of 4,001 reference-solution examples spanning math, science QA, and coding (Table~\ref{tab:mixed_domain_data}). We train a larger Qwen3-8B (with thinking disabled) policy with a smaller Qwen3-4B-Instruct judge. Table~\ref{tab:mixed_domain_8b} shows that \methodname{}-Outcome improves the 8B base policy on every pass@1 evaluation, including math, science, and coding. On the domain-level aggregates, \methodname{}-Outcome is also the strongest Stage-I method on both Math-Aggregate and STEM-Aggregate, for both pass@1 and pass@16. This suggests that reference-guided RL priming is not only learning math-specific templates, but can improve coverage across a broader reasoning mixture. This result highlights another role of reference solutions in \methodname{}: they make reward generation a scaffolded verification task rather than open-ended solution generation. With problem-matched references, a smaller 4B judge can provide useful dense rewards for a larger 8B policy's on-policy traces, suggesting that the judge need not match the policy's scale as long as it is sufficiently capable and reference-conditioned.

Coding is the main exception to the mixed-domain pattern. \methodname{}-Outcome still improves over the base policy on LiveCodeBench, but sparse GRPO remains stronger. We believe this reflects the reward structure of coding tasks: execution provides an unusually strong domain-specific sparse reward, while reference-guided judging is less naturally suited to assigning partial-progress credit. Unlike math or science reasoning, incomplete code may not compile, and many correct implementations can differ substantially from the reference solution. As a result, reference scaffolds provide a weaker substitute for environment feedback in coding than they do in domains where intermediate reductions can be compared against a reference path. This is consistent with the calibration result in Table~\ref{tab:calibration_lcb}, where no-reference judging is about as reliable as reference-conditioned judging on LiveCodeBench, suggesting that the judge relies mainly on functional correctness rather than reference-solution scaffolding. Thus, the coding result is consistent with our view of \methodname{} as Stage-I RL priming: it can improve the starting policy, but when a strong environment reward is available, downstream RL should use it directly.

\begin{table}[t]
\centering
\small
\setlength{\tabcolsep}{8pt}
\renewcommand{\arraystretch}{1.05}
\begin{tabular}{llrr}
\toprule
\textbf{Dataset} & \textbf{Domain} & \textbf{\# Examples} & \textbf{\% of Training} \\
\midrule
InT & Math & 440 & 11.00 \\
POPE & Math & 1,076 & 26.89 \\
SciKnow-Physics & Science & 474 & 11.85 \\
SciKnow-All & Science & 1,000 & 24.99 \\
LCB v6 & Coding & 1,011 & 25.27 \\
\midrule
Total & -- & 4,001 & 100.00 \\
\bottomrule
\end{tabular}
\caption{\footnotesize{\textbf{Mixed-domain Stage-I data.} The additional 8B experiment uses reference-solution examples from math, science QA, and coding.}}
\label{tab:mixed_domain_data}
\end{table}

\begin{table*}[t]
\centering
\scriptsize
\setlength{\tabcolsep}{2.2pt}
\renewcommand{\arraystretch}{1.04}
\begin{adjustbox}{max width=\textwidth}
\begin{tabular}{lrrrrrrrrrrrrrrrrrr}
\toprule
\textbf{Model}
& \multicolumn{2}{c}{\textbf{AIME-25}}
& \multicolumn{2}{c}{\textbf{AIME-26}}
& \multicolumn{2}{c}{\textbf{HMMT-Nov25}}
& \multicolumn{2}{c}{\textbf{IMO-Answer}}
& \multicolumn{2}{c}{\textbf{Math-Agg.}}
& \multicolumn{2}{c}{\textbf{GPQA}}
& \multicolumn{2}{c}{\textbf{OlympiadPhys.}}
& \multicolumn{2}{c}{\textbf{STEM-Agg.}}
& \multicolumn{2}{c}{\textbf{LCB v5}} \\
\cmidrule(lr){2-3}\cmidrule(lr){4-5}\cmidrule(lr){6-7}\cmidrule(lr){8-9}\cmidrule(lr){10-11}\cmidrule(lr){12-13}\cmidrule(lr){14-15}\cmidrule(lr){16-17}\cmidrule(lr){18-19}
& p@1 & p@16 & p@1 & p@16 & p@1 & p@16 & p@1 & p@16 & p@1 & p@16 & p@1 & p@16 & p@1 & p@16 & p@1 & p@16 & p@1 & p@4 \\
\midrule
Qwen3-8B & 19.42 & 43.09 & 16.56 & 45.31 & 10.18 & 38.64 & 15.28 & 40.17 & 15.36 & 41.80 & 47.64 & 88.78 & 35.86 & 64.09 & 41.75 & 76.44 & 36.52 & 43.76 \\
GRPO & 27.14 & 51.58 & 27.71 & 61.83 & 22.38 & 49.61 & 22.51 & 47.42 & 24.93 & 52.61 & 53.41 & 84.41 & 40.36 & 68.18 & 46.88 & 76.30 & \textbf{54.97} & \textbf{64.09} \\
SFT & 5.67 & 29.23 & 5.22 & 27.83 & 3.44 & 21.27 & 5.03 & 29.64 & 4.84 & 26.99 & 37.76 & \textbf{90.66} & 16.45 & 49.56 & 27.11 & 70.11 & 25.66 & 38.28 \\
Self-Distillation & 21.76 & 45.08 & 18.80 & 49.49 & 12.51 & 40.35 & 16.45 & 40.89 & 17.38 & 43.96 & 48.57 & 87.06 & 36.86 & 66.07 & 42.71 & 76.57 & 43.02 & 56.51 \\
\methodname{}-Outcome & \textbf{34.23} & \textbf{58.99} & \textbf{40.90} & \textbf{64.15} & \textbf{25.31} & 47.95 & \textbf{23.36} & 44.25 & \textbf{30.95} & \textbf{53.84} & \textbf{53.46} & 85.31 & \textbf{44.25} & \textbf{68.66} & \textbf{48.86} & \textbf{76.99} & 41.92 & 48.82 \\
\bottomrule
\end{tabular}
\end{adjustbox}
\vspace{-0.2cm}
\caption{\footnotesize{\textbf{8B policy + 4B judge Stage-I results.} All methods are evaluated at 270 steps. \methodname{}-Outcome improves the 8B base model on every pass@1 evaluation and gives the best aggregated results in Math and STEM domains among Stage-I methods.}}
\label{tab:mixed_domain_8b}
\end{table*}

\textbf{Reference and judge calibration.}
We next test whether the dense reward signal actually depends on the problem-matched reference solution, rather than simply reflecting generic LLM-judge confidence or surface-level plausibility. To separate these possibilities, we hold sampled rollouts fixed and vary both judge size and reference condition: a correct problem-matched reference, no reference, or a wrong reference from another problem. We measure misplacement rate, defined as $(\mathrm{FPR}+\mathrm{FNR})/2$, where false positives are incorrect rollouts assigned score $>3$, and false negatives are correct rollouts assigned score $<4$. Lower is better.

Table~\ref{tab:calibration_math_sciknow} shows that, for all 4B-and-larger judges, correct-reference judging gives the lowest misplacement rate across Math, SciKnow-MCQ, and SciKnow-OE. Removing the reference weakens discrimination, and using a wrong reference often makes the reward signal unreliable. The 0.6B judge is unstable, so \methodname{} requires a minimally capable judge. However, the 8B-policy experiment above shows that the judge need not be as large as the policy. Together, these results support the view that the useful reward signal is not generic judge confidence, but verification against a correct problem-matched reference.

Table~\ref{tab:calibration_lcb} shows a different pattern on LiveCodeBench: correct-reference, no-reference, and wrong-reference judging all have similarly low misplacement rates, with no-reference slightly best. This suggests that the coding judge relies more on inferred functional correctness from the code and problem specification than on reference-solution scaffolding. This helps explain why \methodname{}-Outcome improves the base policy on coding, while execution-based sparse GRPO remains especially strong.

\begin{table}[t]
\centering
\small
\setlength{\tabcolsep}{5pt}
\renewcommand{\arraystretch}{1.05}
\begin{tabular}{llrrr}
\toprule
\textbf{LLM Judge} & \textbf{Reference condition} & \textbf{Math} & \textbf{SciKnow-MCQ} & \textbf{SciKnow-OE} \\
\midrule
Qwen3-0.6B & Correct reference & 48.6 & \textbf{42.0} & \textbf{49.4} \\
Qwen3-0.6B & No reference & 48.5 & 47.1 & 50.0 \\
Qwen3-0.6B & Wrong reference & \textbf{47.5} & 44.1 & 51.6 \\
\midrule
Qwen3-4B & Correct reference & \textbf{17.8} & \textbf{14.0} & \textbf{11.4} \\
Qwen3-4B & No reference & 39.2 & 37.5 & 25.2 \\
Qwen3-4B & Wrong reference & 50.4 & 46.0 & 36.7 \\
\midrule
Qwen3-8B & Correct reference & \textbf{18.8} & \textbf{9.8} & \textbf{19.1} \\
Qwen3-8B & No reference & 36.0 & 31.9 & 37.0 \\
Qwen3-8B & Wrong reference & 52.6 & 48.8 & 43.1 \\
\midrule
Qwen3-14B & Correct reference & \textbf{18.2} & \textbf{14.7} & \textbf{12.3} \\
Qwen3-14B & No reference & 38.5 & 29.3 & 27.6 \\
Qwen3-14B & Wrong reference & 50.2 & 47.6 & 36.4 \\
\bottomrule
\end{tabular}
\caption{\footnotesize{\textbf{Calibration misplacement rates on Math and SciKnow.} No-reference judging is weaker, and wrong-reference judging often makes the reward signal unreliable consistently across the 4B, 8B and 14B judges. The 0.6B judge is unstable, which suggests ExpRL does require a minimally capable judge and reliable problem-matched references.}}
\label{tab:calibration_math_sciknow}
\end{table}

\begin{table}[t]
\centering
\small
\setlength{\tabcolsep}{8pt}
\renewcommand{\arraystretch}{1.05}
\begin{tabular}{llr}
\toprule
\textbf{LLM Judge} & \textbf{Reference condition} & \textbf{LiveCodeBench} \\
\midrule
Qwen3-4B & Correct reference & 9.7 \\
Qwen3-4B & No reference & \textbf{8.2} \\
Qwen3-4B & Wrong reference & 10.0 \\
\bottomrule
\end{tabular}
\caption{\footnotesize{\textbf{Calibration misplacement rates on LiveCodeBench.} For coding, problem-matched reference solutions are not critical as the judge primarily depends on tracing the code with input-output pairs.}}
\label{tab:calibration_lcb}
\end{table}

\vspace{-0.3cm}
\section{Related Work and Discussion}
\label{sec:relwork}
\vspace{-0.2cm}

\textbf{Mid-training before RL.}
In modern LLM pipelines, there are broadly two ways to prime a model before an RL run, which we call mid-training. First, \emph{skill-inducing} mid-training imbues useful reasoning behaviors, such as self-correction, backtracking, and verification, useful for exploration and further amplified by RL~\citep{gandhi2025cognitivebehaviorsenableselfimproving,wang2025octothinkermidtrainingincentivizesreinforcement,setlur2025e3learningexploreenables}. Second, \emph{coverage-building} mid-training aims to increase coverage over productive reasoning paths on hard math problems, needed for hard downstream tasks with sparse outcome rewards. Recent pipelines rely on such intermediate stages before RL~\citep{xu2025phi,wang2025octothinkermidtrainingincentivizesreinforcement,su2025scaling}, and recent work studies this interplay explicitly~\citep{zhang2025interplaypretrainingmidtrainingrl}. Traditionally, this coverage is built with SFT on curated traces or rejection-sampled solutions~\citep{zelikman2022star}, but this can narrow the model's exploration during RL~\citep{qu2025pope} and training a model on offline data can cause optimization instabilities~\citep{setlur2026reuse}. \methodname{} focuses on the second setting of mid-training and instead of only cloning traces with correct final answers, it uses \emph{on-policy} RL to reward useful reasoning behaviors and partial progress relative to references. Concurrent work also explores RL for reasoning traces during mid-training stages for improving generation quality~\citep{tan2026self}; our work focuses specifically on building coverage for hard reasoning problems where sparse rewards provide little signal.
\vspace{-0.1cm}

\textbf{Exploration bottleneck in LLM RL.}
Sparse-reward RL is attractive because correctness is often automatically verifiable, but it becomes brittle on hard problems where correct rollouts are rare, leading to under-exploration and sometimes degraded pass@$k$ after RL~\citep{yue2025doesreinforcementlearningreally,zhao2025echochamberrlposttraining}. Prior work improves the learning signal with intrinsic bonuses, entropy regularization, count-based rewards, pass@$n$-aware objectives, and verification-based signals~\citep{gao2025navigateunknownenhancingllm,wang2025reinforcementlearningreasoninglarge,song2025outcomebasedexplorationllmreasoning,chow2024inference,balashankar2025infaligninferenceawarelanguagemodel,zhou2505reinforcing}, as well as by studying the role of earlier training stages~\citep{zhang2025interplaypretrainingmidtrainingrl}. Our approach instead shifts exploration into mid-training, where dense reference-guided rewards can reinforce productive reasoning paths before sparse-reward RL.

\vspace{-0.3cm}
\section{Conclusion}
\vspace{-0.2cm}

We studied RL priming for LLM reasoning through the lens of coverage over productive reasoning paths. \methodname{} uses on-policy RL with dense reference-guided rewards to build this coverage before sparse outcome-reward RL, rewarding partial progress rather than only final correctness. Across answer-based math reasoning benchmarks, this yields a stronger RL-ready initialization and improves downstream sparse-reward RL. Several directions could be further explored. First, \methodname{} currently uses the judge only to produce scalar process rewards; future work could also train policies from the judge’s natural-language feedback, giving the model richer information about which reasoning steps are missing, incorrect, or worth continuing. Second, \methodname{} could be combined with prefix-conditioned generation during post-training, where diverse reasoning prefixes are deliberately sampled or constructed to expand coverage over productive solution paths and better characterize the method’s exploration ceiling. Third, although we explored initial strategies for reducing bias in process rewards, a more systematic study of reward calibration, length normalization, and judge design could help preserve training stability while avoiding excessive length growth.

\textbf{Limitations.}
ExpRL requires auxiliary information, such as reference solutions, to identify the presence of useful techniques and partial progress during mid-training. This may not always be available, especially in domains where good references are hard to obtain.

\vspace{-0.2cm}
\section*{Acknowledgements}
\vspace{-0.2cm}

We thank Anikait Singh, Konwoo Kim and Matthew Yang for feedback, discussions, and help with infrastructure. Violet Xiang was supported by Stanford HAI Hoffman-Yee grants program. Amrith Setlur was supported by a JP Morgan PhD fellowship. Aviral Kumar was supported in part by the Schmidt Sciences AI2050 Early-Career Fellowship. We thank Rogo, Stanford HAI Hoffman-Yee grants program, TPU research cloud, Amazon AWS, and NCSA Delta for providing compute resources that supported this work. 

\bibliography{main}

\newpage
\appendix
\onecolumn

\section{Appendix}
\subsection{Additional Implementation Details}
\label{sec:additional_imp}
We implement the \textbf{Stage-I} experiments by modifying \texttt{verl}\footnote{https://github.com/verl-project/verl} to support on-policy sampling and learning, with up to one-step off-policy updates. The only exception is \textsc{\methodname{}-Process}, which we found performs better with fully on-policy updates. For optimization, we set the upper clipping threshold to \texttt{0.28} for {\methodname{}-Outcome} runs and to \texttt{0.26} for sparse-reward RL (GRPO) runs.

The \textbf{Stage-II} experiments are supported by asynchronous RL pipelines, specifically \texttt{Pipeline-RL}\footnote{https://github.com/ServiceNow/PipelineRL}. We use asynchronous RL for Stage-II primarily to speed up experimentation and to support the large number of training steps required in this stage (500 steps for the answer-based). Each run is performed on a single node with $8$ NVIDIA H100 GPUs.

\subsubsection{Slicing steps for process rewards}
\label{sec:slice_steps}
\begin{figure*}[h]
  \centering
  \includegraphics[width=0.97\textwidth]{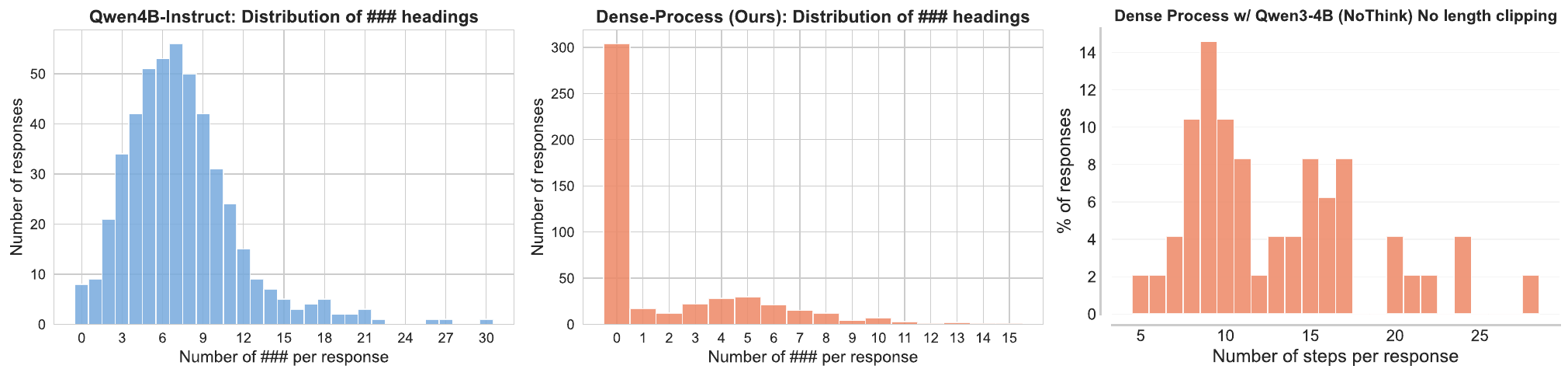}
  \caption{\footnotesize{\textbf{Distribution of \texttt{\#\#\#} step delimiters before and after \methodname{}-Process training.} \textbf{Left:} the base Qwen3-4B-Instruct model naturally emits multiple \texttt{\#\#\#} delimiters per response, which we use to define semantically meaningful prefixes for process rewards. \textbf{Middle:} after \methodname{}-Process training with length clipping, the distribution shifts sharply toward very few delimiters, with many responses containing none or only one. \textbf{Right:} when \methodname{}-Process training is run on Qwen3-4B NoThink without length clipping, the step-count distribution remains much broader. This suggests that the collapse in delimiter count is primarily a side effect of length clipping rather than an inherent consequence of process-level rewards.}}
  \label{fig:step_delim_dist}
\end{figure*}

In Section~\ref{sec:our_method}, we mentioned that we use \texttt{\#\#\#} as the step delimiter when constructing prefixes for process rewards. The main reason is practical: the base model, Qwen3-4B-Instruct, already uses this delimiter by default, so it provides a natural way to break long reasoning traces into semantically meaningful steps. Empirically, about 98.3\% of base-model rollouts contain at least one such delimiter, making it a convenient and robust heuristic for prefix construction. Figure~\ref{fig:step_delim_dist} illustrates the resulting distribution. In the base model (left), responses typically contain multiple \texttt{\#\#\#} headings. However, after \methodname{}-Process training (middle), the distribution shifts sharply toward very few delimiters, and a large fraction of rollouts contain only one step or none at all.

We find that this behavior is largely an artifact of length clipping rather than a fundamental property of process-level rewards. As shown in Figure~\ref{fig:step_delim_dist} (right), when we train a Qwen3-4B hybrid NoThink model with \methodname{}-Process rewards \emph{without} length clipping, the step-count distribution remains much broader and does not collapse in the same way. This suggests that the reduction in the number of \texttt{\#\#\#} delimiters is driven primarily by the interaction between process-level training and the clipped-length penalty. We ultimately do not use this NoThink model in the main experiments because Qwen3-4B-Instruct is more capable and better suited to our study. Nevertheless, the interaction between process rewards, delimiter usage, and length clipping remains an open implementation issue that may be worth revisiting in future work.

\subsubsection{LLM Judge Prompts}
\label{sec:a_prompts}

\begin{figure*}[t]
\scriptsize

\noindent
\begin{minipage}[t]{0.495\textwidth}
\begin{tcolorbox}[
  colback=gray!3,
  colframe=gray!50,
  title={\textbf{Outcome Reward}},
  fonttitle=\scriptsize\bfseries,
  boxrule=0.4pt,
  left=4pt,right=4pt,top=3pt,bottom=3pt,
  boxsep=1.2pt
]
\textbf{Role.} You are an expert mathematician and a meticulous AI reasoning evaluator.

\textbf{Task.} Assess the quality of a \textquotedblleft Generated Reasoning\textquotedblright\ trace.

\textbf{Goal.} Judge how likely it is that a model, after producing the \textquotedblleft Generated Reasoning,\textquotedblright\ would then produce the \emph{exact} \textquotedblleft Reference Solution\textquotedblright\ provided.

You are evaluating the \textbf{logical and causal link} between the reasoning trace and the final answer.

\textbf{Instructions.} First, provide a step-by-step analysis of the connection between the \textquotedblleft Generated Reasoning\textquotedblright\ and the \textquotedblleft Reference Solution.\textquotedblright\ Consider:

\begin{itemize}[nosep,leftmargin=1em,topsep=1pt,partopsep=0pt,itemsep=0pt,parsep=0pt]
\item \textbf{Correctness \& Alignment:} Is the reasoning trace mathematically sound, and does its conclusion \emph{perfectly match} the \textquotedblleft Reference Solution\textquotedblright?
\item \textbf{Sufficiency:} Does the reasoning provide the necessary steps to reach the answer, or does it require a further leap?
\item \textbf{Contradiction:} Does any part of the reasoning contradict the \textquotedblleft Reference Solution\textquotedblright\ or suggest a different answer?
\end{itemize}

After your analysis, provide a single score on the 5-point Likert scale below.

\textbf{Likert scale}
\begin{itemize}[nosep,leftmargin=1em,topsep=1pt,partopsep=0pt,itemsep=0pt,parsep=0pt]
\item \textbf{1 (Very Unlikely):} Incorrect, far short, or leads to a different answer.
\item \textbf{2 (Unlikely):} Significant gaps, flaws, or vagueness.
\item \textbf{3 (Neutral/Possible):} On the right track but incomplete or mildly flawed.
\item \textbf{4 (Likely):} Correct and strongly supports the answer, with only trivial gaps.
\item \textbf{5 (Very Likely):} Sound, complete, and directly implies the reference solution.
\end{itemize}

\textbf{Output format}
\begin{quote}
\textbf{Reasoning:} [Your detailed analysis goes here.]

\textbf{Score:} [1, 2, 3, 4, or 5.]
\end{quote}
\end{tcolorbox}
\end{minipage}\hfill
\begin{minipage}[t]{0.495\textwidth}
\begin{tcolorbox}[
  colback=gray!3,
  colframe=gray!50,
  title={\textbf{Process Reward}},
  fonttitle=\scriptsize\bfseries,
  boxrule=0.4pt,
  left=4pt,right=4pt,top=3pt,bottom=3pt,
  boxsep=1.2pt
]
\textbf{Role.} You are an expert mathematician and a strict AI reasoning evaluator.

\textbf{Inputs.} You will be given:
\begin{itemize}[nosep,leftmargin=1em,topsep=1pt,partopsep=0pt,itemsep=0pt,parsep=0pt]
\item a Math Problem,
\item a Generated Reasoning STEP,
\item a full Reference Solution.
\end{itemize}

\textbf{Task.} Judge how likely it is that this single step/segment could serve as a correct and useful \emph{next step} on a path that matches the Reference Solution and leads to the Reference final answer.

\textbf{Critical constraints}
\begin{itemize}[nosep,leftmargin=1em,topsep=1pt,partopsep=0pt,itemsep=0pt,parsep=0pt]
\item Do not solve the problem yourself.
\item Do not infer missing context or fill in omitted earlier steps.
\item Treat the Reference Solution as ground truth and primary scaffold.
\item If the step uses a different approach than the Reference, count it as aligned only if it explicitly supports the same necessary intermediate claims.
\end{itemize}

\textbf{Reference-conditioned evaluation}
\begin{enumerate}[nosep,leftmargin=1.2em,topsep=1pt,partopsep=0pt,itemsep=0pt,parsep=0pt]
\item Extract 2--6 key checkpoints from the Reference Solution.
\item Determine which checkpoint(s) this step directly establishes, partially supports, is irrelevant to, or contradicts.
\item Decide whether the step is a productive move toward the Reference path.
\end{enumerate}

\textbf{Scoring (5-point Likert)}
\begin{itemize}[nosep,leftmargin=1em,topsep=1pt,partopsep=0pt,itemsep=0pt,parsep=0pt]
\item \textbf{5} = Directly advances the Reference path.
\item \textbf{4} = Likely useful and aligned.
\item \textbf{3} = Possibly useful but weak.
\item \textbf{2} = Unlikely.
\item \textbf{1} = Misleading or contradictory.
\end{itemize}

\textbf{Output format}
\begin{quote}
\textbf{Reasoning:} [Checkpoint-based analysis of what this step supports/contradicts.]

\textbf{Score:} [1$|$2$|$3$|$4$|$5]
\end{quote}
\end{tcolorbox}
\end{minipage}

\caption{System prompts used for LLM-as-judge reward modeling. The left prompt scores full reasoning traces for outcome reward, while the right prompt scores individual reasoning steps for process reward.}
\label{fig:reward-prompts}
\end{figure*}

We use two judge prompts during ExpRL training, corresponding to the two dense reward variants in Section~\ref{sec:our_method}. The first prompt is used for \emph{\methodname{}-Outcome} rewards and scores a full generated reasoning trace against a reference solution. The second prompt is used for \emph{\methodname{}-Process} rewards and scores a single intermediate step or segment against the reference solution, with the goal of providing more localized credit assignment. In both cases, the judge is instructed to \emph{verify} rather than \emph{solve}: it should assess alignment with the reference solution without filling in missing reasoning or correcting the model's mistakes.

\subsection{Ablation: advantage centering for process rewards}
\label{sec:advantage_ablation}
\begin{figure*}
  \centering
  \includegraphics[width=0.95\textwidth]{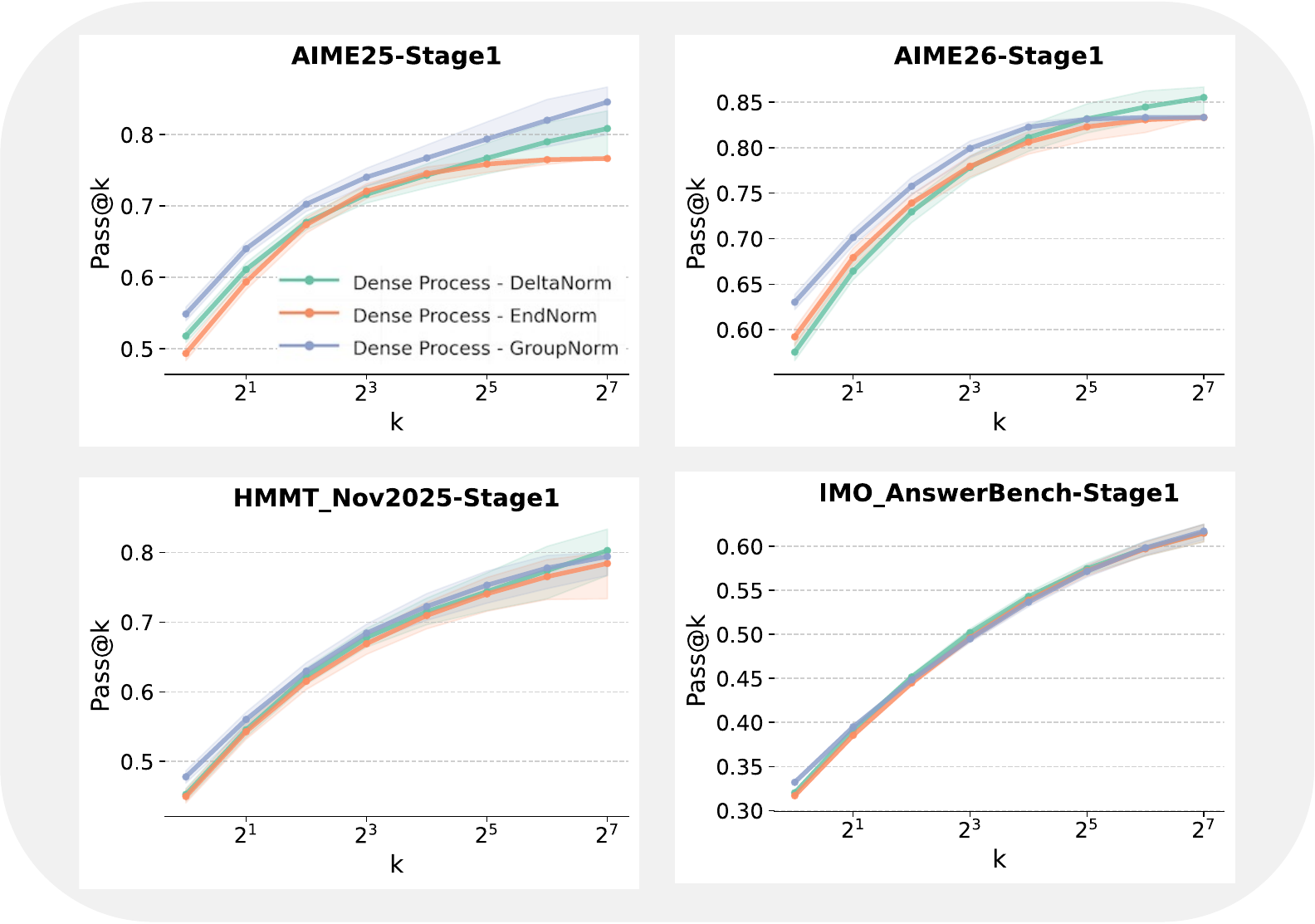}
  \caption{Different strategies for normalizing process rewards.}
  \label{fig:diff_advantage_norms}
\end{figure*}

\vspace{-0.1cm}
We consider several advantage normalizations based on $\{s_t\}$ that emphasize different aspects of partial progress, such as relative improvement over final outcome, local step-to-step gains, or trajectory-centered normalization. These variants operate on the same underlying signal and are described below.
\begin{equation}
A^{\text{EndNorm}}_t(x,y)=
\begin{cases}
s_t - s_T, & \text{if } t < T,\\
s_t,       & \text{if } t = T.
\end{cases}
\label{eq:pr_end}
\end{equation}
\begin{equation}
A^{\text{DeltaNorm}}_t(x,y)=
\begin{cases}
s_t - s_{t-1}, & \text{if } t > 1,\\
s_t - s_T,     & \text{if } t = 1.
\end{cases}
\label{eq:pr_delta_app}
\end{equation}
\begin{equation}
A^{\text{GroupNorm}}_t(x,y)
= s_t - \frac{1}{K}\sum_{k}\frac{1}{T}\sum_{t} s_{t}.
\label{eq:pr_gppo}
\end{equation}
Figure~\ref{fig:diff_advantage_norms} presents the results from three ways of converting prefix scores $\{s_t\}_{t=1}^T$ into segment-level advantages (\methodname{}-Process EndNorm, \methodname{}-Process DeltaNorm, and \methodname{}-Process GroupNorm; Eq.~\ref{eq:pr_end}-\ref{eq:pr_gppo}). Across held-out benchmarks, DeltaNorm, EndNorm, and GroupNorm yield broadly similar Stage-I pass@k curves, indicating that ExpRL is not highly sensitive to the exact centering scheme for process rewards. While GroupNorm is slightly stronger at low k on some benchmarks, these differences are modest and not uniform across tasks.

\subsection{Stage-I Behavior Analysis}
\label{sec:stage1_behavior}

To better understand how RL priming changes the model's reasoning, we perform an LLM-based behavior analysis of the Stage-I rollouts. For each rollout, we send the full model response to an external annotator model, Claude Sonnet 4, and ask it to classify the reasoning using a detailed rubric Figure~\ref{fig:answer-rubric}. The annotator is not asked to solve the problem itself; instead, it reads the generated solution and returns a structured JSON annotation describing which strategies and reasoning behaviors are present.

Our rubric has two aspects: \emph{solution archetypes}, i.e., the high-level strategy used by the rollout, such as coordinatization, casework, recursion, or contradiction, and \emph{reasoning behaviors}, i.e., process-level phenomena such as verification, backtracking, self-correction, exploration, or restart behavior. Most of these are also represented as binary indicators, except for \texttt{self\_interruption} and \texttt{self\_correction}, which are recorded as integer counts. These labels are represented as binary indicators, and multiple archetypes may be active for a single solution if the reasoning combines several strategies.

This analysis is fully LLM-judged: we do not use regexes, keyword matching, or hand-written heuristics. Instead, the annotator applies the rubric definitions directly to the solution text and produces one annotation per rollout. These rollout-level annotations are then grouped by problem and aggregated by downstream analysis scripts. In particular, for each problem we average the annotations across rollouts from the same model, compare them against the corresponding statistics for the base Qwen3-4B-Instruct model, and then count how many problems exhibit gains or losses in each behavior. This allows us to quantify not only which behaviors are present, but how RL priming changes the distribution of behaviors relative to the base model.

\begin{figure*}[t]
\small
\begin{tcolorbox}[
  colback=gray!3,
  colframe=gray!50,
  title={\textbf{System Prompt: Answer-Problem Annotation Rubric}},
  fonttitle=\small\bfseries,
  boxrule=0.5pt,
  left=6pt, right=6pt, top=4pt, bottom=4pt
]

\textbf{Task.} Classify the given solution along two dimensions: (1) solution archetypes and (2) reasoning behaviors. Output only a JSON object.

\medskip
\textbf{\textsc{Layer 1: Solution Archetypes}} {\scriptsize (all 0/1)}

\begin{itemize}[nosep,leftmargin=1em]
\item \texttt{coordinate\_bash} --- Places a geometric figure on a coordinate system and reduces geometry to algebraic equations. \textit{Skip if coordinates are given in the problem or appear only briefly.}
\item \texttt{algebraic\_system\_solving} --- Main work is manipulating/solving a system of equations: expanding, subtracting, factoring. \textit{Skip if equations arise from coordinates (that's coordinate\_bash) or from symmetric structure.}
\item \texttt{symmetric\_function\_reduction} --- Recognizes symmetric/cyclic structure; uses Vieta's, elementary symmetric polynomials, or WLOG. \textit{Skip if the solution merely uses the quadratic formula.}
\item \texttt{casework} --- Splits into explicitly labeled cases based on a key parameter. \textit{Skip if only testing a few small values (that's constructive\_witness).}
\item \texttt{constructive\_witness\_and\_bound} --- Two phases: (1) construct an explicit example achieving a target, (2) prove no better value is possible. \textit{Skip if only one phase is present.}
\item \texttt{inclusion\_exclusion\_partition} --- Partitions a set into disjoint regions or applies inclusion-exclusion to count overlapping sets.
\item \texttt{expectation\_decomposition} --- Decomposes a count/expected value into indicator-variable contributions via linearity of expectation.
\item \texttt{recursion\_dp} --- Sets up a recurrence relation or DP table and builds up from base cases.
\item \texttt{number\_theoretic\_analysis} --- Main approach involves divisibility, modular arithmetic, Euler's totient, or CRT. \textit{Skip if mod arithmetic is used only briefly.}
\item \texttt{digit\_column\_analysis} --- Analyzes a cryptarithmetic problem column by column, tracking carries.
\item \texttt{geometric\_transformation} --- Applies rotation, reflection, inversion, or complex multiplication as the key insight. \textit{Skip if the solution just sets up coordinates.}
\item \texttt{parity\_invariant} --- Uses parity, coloring, monovariant, or invariant as a key structural insight. \textit{Skip if even/odd is a minor observation.}
\end{itemize}

\medskip
\textbf{\textsc{Layer 2: Reasoning Behaviors}}

\begin{itemize}[nosep,leftmargin=1em]
\item \texttt{self\_interruption} {\scriptsize (count)} --- Model halts mid-thought to flag a problem (``Wait --- this contradicts...''). \textit{Skip narrative ``we wait'' or ``wait'' in problem text.}
\item \texttt{self\_correction} {\scriptsize (count)} --- Model explicitly revises a prior claim (``Actually, the above is incorrect...''). \textit{Skip filler ``actually'' without a specific claim being corrected.}
\item \texttt{error\_acknowledgment} {\scriptsize (0/1)} --- Model admits its OWN reasoning was wrong. \textit{Skip ``this case is impossible'' when eliminating a mathematical possibility.}
\item \texttt{exploration} {\scriptsize (0/1)} --- Model proposes a meaningfully different strategy. \textit{Skip routine next steps within the same approach.}
\item \texttt{redo\_restart} {\scriptsize (0/1)} --- Model abandons significant prior work and starts over. \textit{Skip reformatting or restating.}
\item \texttt{hedging} {\scriptsize (0/1)} --- Model expresses genuine uncertainty (``I think this might work, but I'm not sure''). \textit{Skip ``I think of this as a graph problem'' (framing, not doubt).}
\item \texttt{backtracking} {\scriptsize (0/1)} --- Model returns to an earlier decision point to try a different branch. \textit{Skip referencing earlier work without changing direction.}
\item \texttt{structured\_steps} {\scriptsize (0/1)} --- Solution has explicitly labeled sequential stages (\texttt{\#\#\#}, \texttt{**Step N**}).
\item \texttt{verification} {\scriptsize (0/1)} --- Model plugs its answer back into the original problem AFTER deriving it. \textit{Skip forward-reasoning ``check whether $f$ is injective.''}
\end{itemize}

\medskip
\textbf{Output format.} Respond with ONLY a JSON object containing all 21 fields above. No explanation.
\end{tcolorbox}
\caption{System prompt used for LLM-as-judge annotation of answer-based competition mathematics solutions. Each solution is annotated for 12 solution archetypes (Layer~1) and 9 reasoning behaviors (Layer~2). Italicized text indicates disambiguation rules to prevent common false positives.}
\label{fig:answer-rubric}
\end{figure*}

\subsection{Stage-I pass@k curves on held-out benchmarks}
\label{sec:pass@k}

\begin{figure*}[h]
  \centering
  \includegraphics[width=0.97\textwidth]{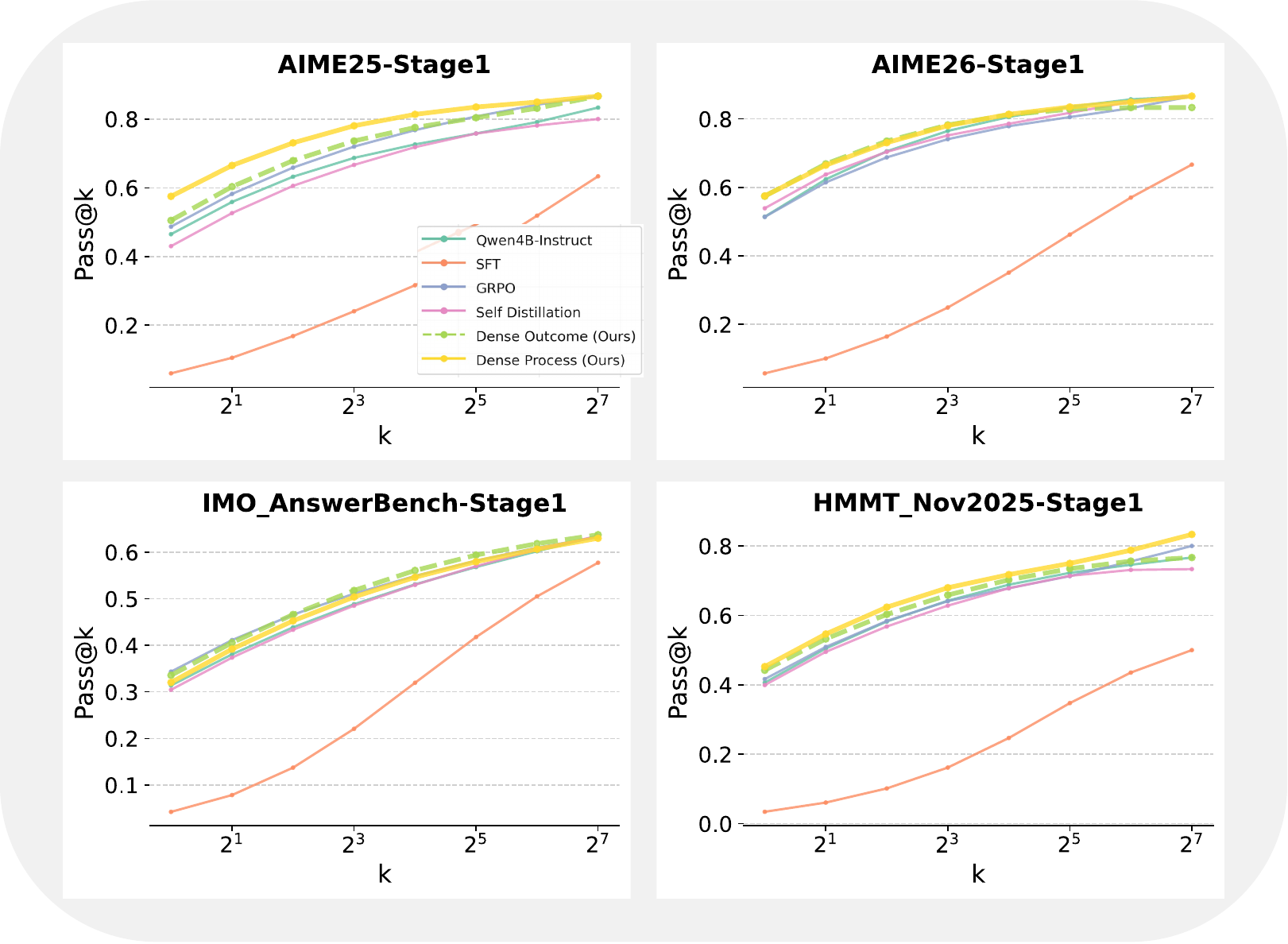}
  \caption{
Stage-I pass@k on held-out answer-based benchmarks after RL priming. \methodname{} improves sampling efficiency prior to subsequent sparse-reward RL, with the clearest gains appearing at low to moderate \(k\), where finding a correct solution in only a few attempts is most difficult. \methodname{}-Outcome and \methodname{}-Process rewards consistently outperform imitation-based and sparse-reward-only baselines on AIME25, AIME26, and HMMT, while gains on IMO-AnswerBench are smaller.}
  \label{fig:stage1_passk}
\end{figure*}

Figure~\ref{fig:stage1_passk} shows that \methodname{} improves pass@k on held-out answer-based benchmarks immediately after RL priming, before any downstream sparse-reward RL. The effect is strongest on AIME25, AIME26, and HMMT, especially at low to moderate \(k\), where finding a correct rollout is hardest. Gains on IMO-AnswerBench are smaller, suggesting that the advantage of RL priming is most visible on the harder held-out tasks. Taken together, these results indicate that \methodname{} already yields a better RL-ready initialization at Stage-I by improving useful coverage under sampling.

\subsection{The LLM judge provides a useful dense learning signal}
\begin{figure*}[h]
  \centering
  \includegraphics[width=0.97\textwidth]{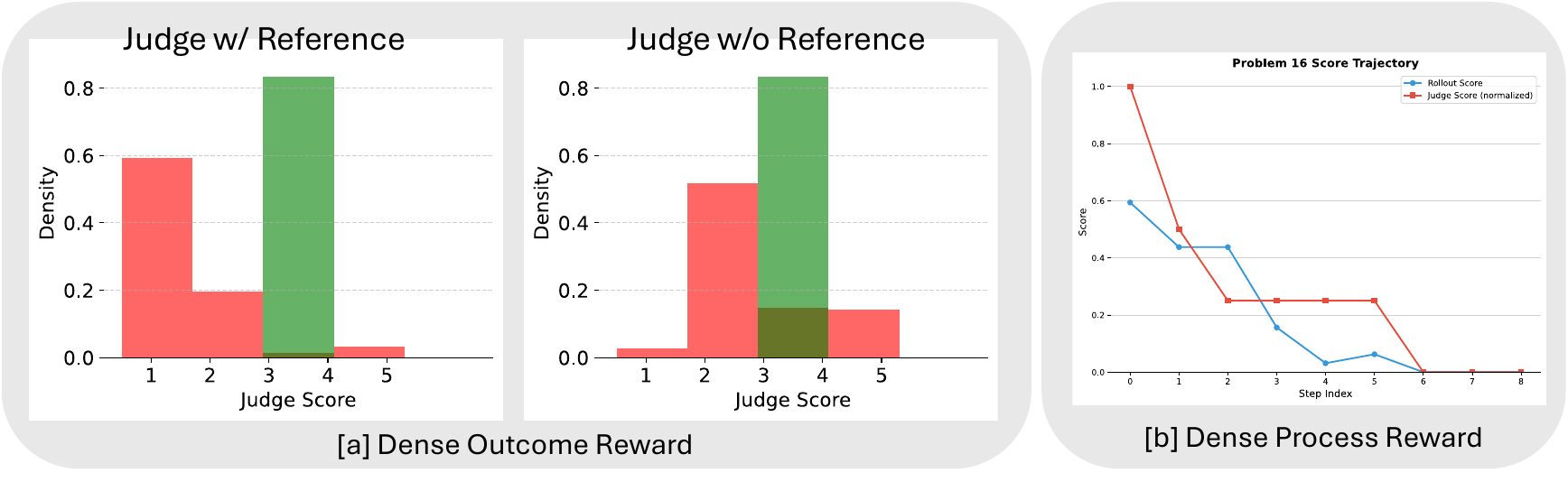}
  \caption{LLM judge score calibration on the RL priming set.
    [a] ExpRL-Outcome reward. Red bars indicate incorrect final answers and green bars indicate correct final answers. Solid bars use the reference solution; dashed bars omit the reference solution.
    [b] ExpRL-Process reward. For each prefix, we estimate downstream success by sampling 32 continuations and compare this success trajectory to the judge-score trajectory. Process scores are noisier than outcome scores but broadly track eventual success.}
  \label{fig:reward_calibration}
\end{figure*}

A core premise of \methodname{} is that a reference-conditioned judge can provide informative dense feedback even when fully correct solutions are rare. Figure~\ref{fig:reward_calibration} supports this premise at both the outcome and process levels. At the outcome level, correct rollouts receive substantially higher judge scores than incorrect ones, with clearer separation when the judge is conditioned on the reference solution. At the process level, prefix scores are noisier but still broadly track how likely a prefix is to lead to eventual success. Together, these results indicate that the judge preserves enough ranking information to distinguish promising intermediate progress from regressions, providing exactly the richer learning signal that sparse-reward RL lacks on hard problems.

\end{document}